\documentclass[journal]{IEEEtran}

\ifCLASSOPTIONonecolumn
	\usepackage{setspace} 
\fi

\usepackage[utf8]{inputenc}
\usepackage{graphicx}
\usepackage{amssymb}
\usepackage{amsmath}
\usepackage{amsfonts}
\usepackage{multirow}
\usepackage{cite}
\usepackage{color}
\usepackage[table]{xcolor}
\usepackage[english]{babel}
\usepackage{algorithm}
\usepackage{algorithmic}
\usepackage{setspace}
\usepackage[caption=false,font=footnotesize]{subfig}
\usepackage{url}
\usepackage{booktabs}
\usepackage{multirow}
\usepackage{flushend}
\usepackage{epstopdf}

\newsavebox{\ieeealgbox}

\newcommand{\vect}[1]{\mathbf{#1}}
\newcommand{\ra}[1]{\renewcommand{\arraystretch}{#1}}

\DeclareMathSymbol{\R}{\mathalpha}{AMSb}{"52}

\newcommand{\norm}[2][2]{\left\lVert #2 \right\rVert_{#1}}

\newcommand{\boldbeta}{\ensuremath{\boldsymbol{\beta}}}

\begin{document}

\title{Group Sparse Regularization for \\ Deep Neural Networks}

\author{Simone~Scardapane,~\IEEEmembership{Student~Member,~IEEE,} %
        Danilo~Comminiello,~\IEEEmembership{Member,~IEEE,} %
        Amir~Hussain,~\IEEEmembership{Member,~IEEE,} %
        and~Aurelio~Uncini,~\IEEEmembership{Member,~IEEE} %
\thanks{Simone Scardapane, Danilo Comminiello and Aurelio Uncini are with the Department of Information Engineering, Electronics and Telecommunications (DIET), ``Sapienza'' University of Rome, Via Eudossiana 18, 00184, Rome. Amir Hussain is with the Division of Computing Science \& Maths, School of Natural Sciences, University of Stirling, Stirling FK9 4LA, Scotland, UK.
Email: \{simone.scardapane,danilo.comminiello,aurelio.uncini\}@uniroma1.it; ahu@cs.stir.ac.uk.}
\thanks{Corresponding Author: Simone Scardapane.}}

\markboth{}%
{Scardapane \MakeLowercase{\textit{et al.}}: Group LASSO for Deep Neural Networks}

\maketitle

\begin{abstract}
In this paper, we consider the joint task of simultaneously optimizing (i) the weights of a deep neural network, (ii) the number of neurons for each hidden layer, and (iii) the subset of active input features (i.e., feature selection). While these problems are generally dealt with separately, we present a simple regularized formulation allowing to solve all three of them in parallel, using standard optimization routines. Specifically, we extend the group Lasso penalty (originated in the linear regression literature) in order to impose group-level sparsity on the network's connections, where each group is defined as the set of outgoing weights from a unit. Depending on the specific case, the weights can be related to an input variable, to a hidden neuron, or to a bias unit, thus performing simultaneously all the aforementioned tasks in order to obtain a compact network. We perform an extensive experimental evaluation, by comparing with classical weight decay and Lasso penalties. We show that a sparse version of the group Lasso penalty is able to achieve competitive performances, while at the same time resulting in extremely compact networks with a smaller number of input features. We evaluate both on a toy dataset for handwritten digit recognition, and on multiple realistic large-scale classification problems.
\end{abstract}


\IEEEpeerreviewmaketitle

\section{Introduction}
\label{sec:introduction}

The recent interest in deep learning has made it feasible to train very deep (and large) neural networks, leading to remarkable accuracies in many high-dimensional problems including image recognition, video tagging, biomedical diagnosis, and others \cite{schmidhuber2015deep,lecun2015deep}. While even five hidden layers were considered challenging until very recently, today simple techniques such as the inclusion of interlayer connections \cite{he2015deep} and dropout \cite{srivastava2014dropout}  allow to train networks with hundreds (or thousands) of hidden layers, amounting to millions (or billions) of adaptable parameters. At the same time, it becomes extremely common to `overpower' the network, by providing it with more flexibility and complexity than strictly required by the data at hand. Arguments that favor simple models instead of complex models for describing a phenomenon are quite known in the machine learning literature \cite{domingos2012few}. However, this is actually far from being just a philosophical problem of `choosing the simplest model'. Having too many weights in a network can clearly increase the risk of overfitting; in addition, their exchange is the main bottleneck in most parallel implementations of gradient descent, where agents must forward them to a centralized parameter server \cite{recht2011hogwild,seide2014parallelizability}; and finally, the resulting models might not work on low-power or embedded devices due to excessive computational power needed for performing dense, large matrix-matrix multiplications \cite{courbariaux2015binaryconnect}.

In practice, current evidence points to the fact that the majority of weights in most deep network are not necessary to its accuracy. As an example, Denil \textit{et al.} \cite{denil2013predicting} demonstrated that it is possible to learn only a small percentage of the weights, while the others can be predicted using a kernel-based estimator, resulting in most cases in a negligible drop in terms of classification accuracy. Similarly, in some cases it is possible to replace the original weight matrix with a low-rank approximation, and perform gradient descent on the factor matrices \cite{sainath2013low}. Driven by these observations, recently the number of works trying to reduce the network's weights have increased drastically. Most of these works either require strong assumptions on the connectivity (e.g, the low-rank assumption), or they require multiple, separate training steps. For example, the popular pruning method of Han \textit{et al.} \cite{han2015learning} works by first training a network, setting to zero all the weights based on a fixed threshold, are then fine-tuning the remaining connections with a second training step. Alternatively, learned weights can be reduced by applying vector quantization techniques \cite{gong2014compressing}, which however are formulated as a separate optimization problem. There are endless other possibilities, e.g. (i) we can use `distillation' to train a separate, smaller network that imitates the original one, as popularized by Hinton \textit{et al.} \cite{hinton2015distilling}; (ii) we can leverage over classical works on pruning, such as the optimal brain damage algorithm, that uses second-order information on the gradient of the cost function to remove `non salient' connections after training \cite{lecun1989optimal}; (iii) we can work with limited numerical precision to reduce storage \cite{gupta2015deep} (up to the extreme of a single bit per weight \cite{courbariaux2015binaryconnect}); or we can use hash functions to force weight sharing \cite{chen2015compressing}; and so on.

When considering high-dimensional datasets, an additional problem is that of feature selection, where we search for a small subset of input features that brings most of the discriminative information \cite{guyon2003introduction}. Feature selection and pruning are related problems: adding a new set of features to a task generally results in the need of increasing the network's capacity (in terms of number of neurons), all the way up to the last hidden layer. Similarly to before, there are countless techniques for feature selection (or dimensionality reduction of the input vector), including  principal component analysis, mutual information \cite{kwak2002input}, autoencoders, and many others. What we obtain, however, is a rather complex workflow of machine learning primitives: one algorithm to select features; an optimization criterion for training the network; and possibly another procedure to compress the weight matrices. This raises the following question, which is the main motivation for this paper: is there a principled way of performing all three tasks \textit{simultaneously}, by minimizing a properly defined cost function? This is further motivated by the fact that, in a neural network, pruning a node and deleting an input feature are almost equivalent problems. In fact, it is customary to consider the input vector as an additional layer of the neural network, having no ingoing connections and having outgoing connections to the first hidden layer. In this sense, pruning a neuron from this initial layer can be considered the same as deleting the corresponding input feature.

Currently, the only principled way to achieve this objective is the use of $\ell_1$ regularization, wherein we penalize the sum of absolute values of the weights during training. The $\ell_1$ norm acts as a convex proxy of the non-convex, non-differentiable $\ell_0$ norm \cite{tibshirani1996regression}. Its  use originated in the linear regression routine, where it is called the Lasso estimator, and it has been widely popularized recently thanks to the interest in compressive sensing \cite{candes2008introduction,bach2012optimization}. Even if it has a non differentiable point in $0$, in practice this rarely causes problems to standard first-order optimizers. In fact, it is common to simultaneously impose both weight-level sparsity with the $\ell_1$ norm, and weight minimization using the $\ell_2$ norm, resulting in the so-called `elastic net' penalization \cite{zou2005regularization}. Despite its popularity, however, the $\ell_1$ norm is only an indirect way of solving the previously mentioned problems: a neuron can be removed if, and only if, all its ingoing or outgoing connections have been set to $0$. In a sense, this is highly sub-optimal: between two equally sparse networks, we would prefer one which has a more \textit{structured} level of sparsity, i.e. with a smaller number of neurons per layer.

In this paper, we show how a simple modification of the Lasso penalty, called the `group Lasso' penalty in the linear regression literature \cite{yuan2006model,schmidt2010graphical}, can be used efficiently to this end. A group Lasso formulation can be used to impose sparsity on a group level, such that all the variables in a group are either simultaneously set to $0$, or none of them are. An additional variation, called the sparse group Lasso, can also be used to impose further sparsity on the non-sparse groups \cite{friedman2010note,simon2013sparse}. Here, we apply this idea by considering all the outgoing weights from a neuron as a single group. In this way, the optimization algorithm is able to remove entire neurons at a time. Depending on the specific neuron, we obtain different effects, corresponding to what we discussed before: feature selection when removing an input neuron; pruning when removing an internal neuron; and also bias selection when considering a bias unit (see next section). The idea of group $\ell_1$ regularization in machine learning is quite known when considering convex loss functions \cite{jenatton2011structured}, including multikernel \cite{bach2008consistency} and multitask problems \cite{liu2009multi}. However, to the best of our knowledge, such a general formulation was never considered in the neural networks literature, except for very specific cases. For example, Zhao \textit{et al.} \cite{zhao2015heterogeneous} used a group sparse penalty to select groups of features co-occurring in a robotic control task. Similarly, Zhu \textit{et al.} \cite{zhu2016co} have used a group sparse formulation to select informative groups of features in a multi-modal context.

On the contrary, in this paper we use the group Lasso formulation as a generic tool for enforcing compact networks with a lower subset of selected features. In fact, our experimental comparisons easily show that the best results are obtained with the sparse group term, where we can obtain comparable accuracies to $\ell_2$-regularized ans $\ell_1$-regularized networks, while at the same time reducing by a large margin the number of neurons in every layer. In addition, the regularizer can be implemented immediately in most existing software libraries, and it does not increase the computational complexity with respect to a traditional weight decay technique.

\subsubsection*{\textbf{Outline of the paper}} The paper is organized as follows. Section \ref{sec:weight_regularization} describes standard techniques for regularizing a neural network during training, namely $\ell_2$, $\ell_1$ and composite $\ell_2$/$\ell_1$ terms. Section \ref{sec:neuron_regularization} describes our novels group Lasso and sparse group Lasso penalties, showing the meaning of groups in this context. Then, we test our algorithms in Section \ref{sec:experimental_results} on a simple toy dataset of handwritten digits recognition, followed by multiple realistic experiments with standard deep learning benchmarks. After going more in depth with respect to some related pruning techniques in Section \ref{sec:related_works}, we conclude with some final remarks in Section \ref{sec:conclusions}.

\subsubsection*{\textbf{Notation}}
In the rest of the paper, vectors are denoted by boldface lowercase letters, e.g. $\vect{a}$, while matrices are denoted by boldface uppercase letters, e.g. $\vect{A}$. All vectors are assumed column vectors. The operator $\norm[p]{\cdot}$ is the standard $\ell_p$ norm on an Euclidean space. For $p=2$ this is the Euclidean norm, while for $p=1$ we obtain the Manhattan (or taxicab) norm defined for a generic vector $\boldbeta \in \R^B$ as $\norm[1]{\boldbeta} = \sum_{k=1}^B |\beta_k|$.

\section{Weight-level regularization for neural networks}
\label{sec:weight_regularization}

Let us denote by $\vect{y} = \vect{f}(\vect{x}; \vect{w})$ a generic deep neural network, taking as input a vector $\vect{x} \in \R^d$, and returning a vector $\vect{y} \in \R^o$ after propagating it through $H$ hidden layers. The vector $\vect{w} \in \R^Q$ is used as a shorthand for the column-vector concatenation of all adaptable parameters of the network. The generic $k$th hidden layer, $1 \le k \le H+1$, operates on a $L_k$-dimensional input vector $\vect{h}_k$ and returns an $L_{k+1}$-dimensional output vector $\vect{h}_{k+1}$ as:
\begin{equation}
\vect{h}_{k+1} = g_k\left( \vect{W}_k \vect{h}_k + \vect{b}_k \right) \,,
\end{equation}
where $\left\{ \vect{W}_k, \vect{b}_k \right\}$ are the adaptable parameters of the layer, while $g_k(\cdot)$ is a properly chosen activation function to be applied element-wise. By convention we have $\vect{h}_1 = \vect{x}$. For training the weights of the network, consider a generic training set of $N$ examples given by $\left\{ \left(\vect{x}_1, \vect{d}_1\right), \ldots, \left(\vect{x}_N, \vect{d}_N\right) \right\}$. The network is trained by minimizing a standard regularized cost function:
\begin{equation}
\vect{w}^* = \underset{\vect{w}}{\arg\min} \left\{ \frac{1}{N} \sum_{i=1}^N L(\vect{d}_i, \vect{f}(\vect{x}_i)) + \lambda R(\vect{w}) \right\} \,,
\label{eq:nn_cost_function}
\end{equation}
where $L(\cdot, \cdot)$ is a proper cost function, $R(\cdot)$ is used to impose regularization, and the scalar coefficient $\lambda \in \R^+$ weights the two terms. Standard choices for $L(\cdot, \cdot)$ are the squared error for regression problems, and the cross-entropy loss for classification problems \cite{haykin2009neural}.

By far the most common choice for regularizing the network, thus avoiding overfitting, is to impose a squared $\ell_2$ norm constraint on the weights:
\begin{equation}
R_{\ell_2}(\vect{w})  \triangleq \norm[2]{\vect{w}}^2 \,.
\label{eq:ell_2_regularization}
\end{equation}
In the neural networks' literature, this is commonly denoted as `weight decay' \cite{moody1995simple}, since in a steepest descent approach, its net effect is to reduce the weights by a factor proportional to their magnitude at every iteration. Sometimes it is also denoted as Tikhonov regularization. However, the only way to enforce sparsity with weight decay is to artificially force to zero all weights that are lower, in absolute terms, than a certain threshold. Even in this way, its sparsity effect might be negligible.

As we stated in the introduction, the second most common approach to regularize the network, inspired by the Lasso algorithm, is to penalize the absolute magnitude of the weights:
\begin{equation}
R_{\ell_1}(\vect{w})  \triangleq \norm[1]{\vect{w}} = \sum_{k=1}^Q |w_k| \,.
\label{eq:ell_1_regularization}
\end{equation}
The $\ell_1$ formulation is not differentiable at $0$, where it is necessary to resort to a subgradient formulation. Everywhere else, its gradient is constant, and in a standard minimization procedure it moves each weight by a constant factor towards zero (in the next section, we also provide a simple geometrical intuition on its behavior). While there exists customized algorithms to solve non-convex $\ell_1$ regularized problems \cite{ochs2015iteratively}, it is common in the neural networks' literature to apply directly the same first-order procedures (e.g., stochastic descent with momentum) as for the weight decay formulation. As an example, all libraries built on top of the popular Theano framework \cite{bergstra2010theano} assigns a default gradient value of $0$ to terms such that $w_k=0$. Due to this, a thresholding step after optimization is generally required also in this case to obtain precisely sparse solutions \cite{bengio2012practical}, although the resulting level of sparsity is quite higher than using weight decay.

One popular variation is to approximate the $\ell_1$ norm by a convex term, e.g. $\norm[1]{\vect{w}} = \sum_{k=1}^Q \sqrt{w_k^2 + \beta}$ for a sufficiently small scalar factor $\beta$, to obtain a smooth problem. Another possibility is to consider a mixture of $\ell_2$ and $\ell_1$ regularization, which is sometimes denoted as elastic net penalization \cite{zou2005regularization}. The problem in this case, however, is that it is required to select two different hyper-parameters for weighting differently the two terms.

\section{Neuron-level regularization with group sparsity}
\label{sec:neuron_regularization}

\subsection{Formulation of the algorithm}

\begin{figure}
\centering
\includegraphics[scale=0.85]{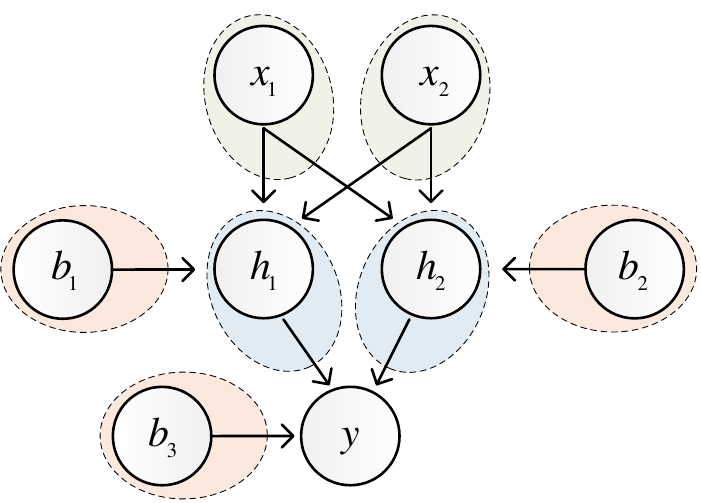}
\caption{Schematic representation of a group LASSO regularization with two inputs (top), two hidden neurons with biases (middle), and one output (bottom). We have three groups of connections. \textbf{Green}: input groups; \textbf{blue}: hidden groups; \textbf{red}: bias groups.}
\label{fig:MLP}
\end{figure}

Both $\ell_2$ regularization in \eqref{eq:ell_2_regularization} and $\ell_1$ regularization in \eqref{eq:ell_1_regularization} are efficient for preventing overfitting, but they are not optimal for obtaining compact networks. Generally speaking, a neuron can be removed from the architecture only if all its connections (either ingoing or outgoing) have been zeroed out during training. However, this objective is not actively pursued while minimizing the cost in \eqref{eq:nn_cost_function}. Between the many local minima, some might be equivalent (or almost equivalent) in terms of accuracy, while corresponding to more compact and efficient networks. As there is no principled way to converge to one instead of the other, when using these kind of regularization the resulting network's design will simply be a matter of initialization of the optimization procedure.

The basic idea of this paper is to consider \textit{group-level} sparsity, in order to force all outgoing connections from a single neuron (corresponding to a group) to be either simultaneously zero, or not. More specifically, we consider three different groups of variables, corresponding to three different effects of the group-level sparsity:
\begin{enumerate}
\item \textbf{Input groups} $\mathcal{G}_{\text{in}}$: a single element $\vect{g_i} \in \mathcal{G}_{\text{in}}, i = 1,\ldots,d$ is the vector of all outgoing connections from the $i$th input neuron to the network, i.e. it corresponds to the first row transposed of the matrix $\vect{W}_1$.
\item \textbf{Hidden groups} $\mathcal{G}_{\text{h}}$: in this case, a single element $\vect{g} \in \mathcal{G}_{\text{h}}$ corresponds to the vector of all outgoing connections from one of the neurons in the hidden layers of the network, i.e. one row (transposed) of a matrix $\vect{W}_k, k > 1$. There are $\sum_{k=2}^{H+1} N_k$ such groups, corresponding to neurons in the internal layers up to the final output one.
\item \textbf{Bias groups} $\mathcal{G}_{\text{b}}$: these are one-dimensional groups (scalars) corresponding to the biases on the network, of which there are $\sum_{k=1}^{H+1} N_k$. They correspond to a single element of the vectors $\left\{\vect{b}_1, \ldots, \vect{b}_{H+1} \right\}$.
\end{enumerate}
Overall, we have a total of $G = 2\sum_{k=1}^{H+1} N_k$ groups, corresponding to three specific effects on the resulting network. If the variables of an input group are set to zero, the corresponding feature can be neglected during the prediction phase, effectively corresponding to a feature selection procedure. Then, if the variables in an hidden group are set to zero, we can remove the corresponding neuron, thereby obtaining a pruning effect and a thinner hidden layer. Finally, if a variable in a bias group is set to zero, we can remove the corresponding bias from the neuron. We note that having a separate group for every bias is not the unique choice. We can consider having a single bias unit for every layer feeding every neuron in that layer. In this case, we would have a single bias group per layer, corresponding to keeping or deleting every bias in it. Generally speaking, we have not found significant improvements in one way or the other.

A visual representation of this weight grouping strategy is shown in Fig. \ref{fig:MLP} for a simple network with two inputs (top of the figure), one hidden layer with two units (middle of the figure), and a single output unit (bottom of the figure). In the figure, input groups are shown with a green background; hidden groups (which in this case have a single element per group) are shown with a blue background; while the $3$ bias groups are surrounded in a light red background.

Let us define for simplicity the total set of groups as 
$$\mathcal{G} = \mathcal{G}_{\text{in}} \cup \mathcal{G}_{\text{h}} \cup \mathcal{G}_{\text{b}} \,.$$
Group sparse regularization can be written as \cite{yuan2006model}:
\begin{equation}
R_{\ell_{2,1}}(\vect{w}) \triangleq \displaystyle\sum_{\vect{g} \in \mathcal{G}} \sqrt{|\vect{g}|} \norm{\vect{g}} \,, 
\label{eq:reg_group_lasso}
\end{equation}
where $|\vect{g}|$ denotes the dimensionality of the vector $\vect{g}$, and it ensures that each group gets weighted uniformly. Note that, for one-dimensional groups, the expression in \eqref{eq:reg_group_lasso} simplifies to the standard Lasso. Similarly to the $\ell_1$ norm, the term in \eqref{eq:reg_group_lasso} is convex but non-smooth, since its gradient is not defined if $\norm{\vect{g}}=0$. The sub-gradient of a single term in  \eqref{eq:reg_group_lasso} is given by:
\begin{equation}
\frac{\partial \left\{ \sqrt{|\vect{g}|} \norm{\vect{g}} \right\}}{\partial\vect{g}} = 
	\begin{cases}
		\sqrt{|\vect{g}|}\displaystyle\frac{\vect{g}}{\norm{\vect{g}}} & \text{if } \vect{g} \neq \vect{0} \\
		\Bigl\{\sqrt{|\vect{g}|}\vect{t} : \norm{\vect{t}} \le 1 \Bigr\} & \text{otherwise}
	\end{cases}
\end{equation}
As for the $\ell_1$ norm, we have found very good convergence behaviors using standard first-order optimizers, with a default choice of $\vect{0}$ as sub-gradient in the second case. Also here, a final thresholding step is required to obtain precisely sparse solutions. Note that we have used the $\ell_{2,1}$ symbol in \eqref{eq:reg_group_lasso} as the formulation is closely related to the $\ell_{2,1}$ norm defined for matrices.

The formulation in \eqref{eq:reg_group_lasso} might still be sub-optimal, however, since we lose guarantees of sparsity at the level of single connections among those remaining after removing some of the groups. To force this, we also consider the following composite `sparse group Lasso' (SGL) penalty \cite{friedman2010note,simon2013sparse}:
\begin{equation}
R_{\text{SGL}}(\vect{w}) \triangleq R_{\ell_{2,1}}(\vect{w}) + R_{\ell_{1}}(\vect{w}) \,.
\label{eq:sgl_regularization}
\end{equation}
The SGL penalty has the same properties as its constituting norms, namely, it is convex but non-differentiable. Differently from an elastic net penalization, we have found that optimal results can be achieved by considering a single regularization factor for both terms in \eqref{eq:sgl_regularization}.

\begin{figure}
\centering
\includegraphics[scale=1]{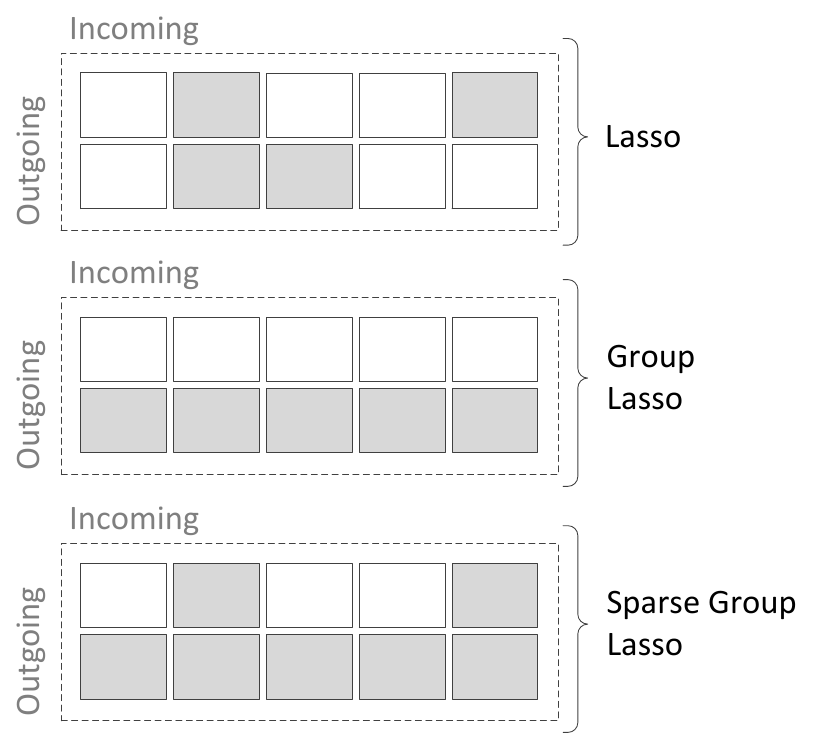}
\caption{Comparison between Lasso, group Lasso, and sparse group Lasso applied to a single weight matrix. In gray we represent the removed connections.}
\label{fig:Comparison_1}
\end{figure}

A visual comparison between $\ell_1$, $\ell_{2,1}$, and SGL penalizations is given in Fig. \ref{fig:Comparison_1}. The dashed box represents one weight matrix connecting a $2$-dimensional input layer to a $5$-dimensional output layer. In gray, we show a possible combination of matrix elements that are zeroed out by the corresponding penalization. The Lasso penalty removes elements without optimizing neuron-level considerations. In this example, we remove $4$ connections (thus obtaining a $40\%$ level of sparsity), and we might remove the second neuron from the second layer (only in case the bias unit to the neuron has also been deleted). The group Lasso penalization removes all connections exiting from the second neuron, which can now be safely removed from the network. The sparsity level is just slightly higher than in the first case, but the resulting connectivity is more structured. Finally, the SGL formulation combines the advantages of both formulation: we remove all connections from the second neuron in the first layer \textit{and} two of the remaining connections, thus achieving a $70\%$ level of sparsity in the layer and an extremely compact (and power-efficient) network.

\subsection{Graphical interpretation of group sparsity}

The group Lasso penalty admits a very interesting geometrical interpretation whenever the first term in \eqref{eq:nn_cost_function} is convex (see for example \cite[Section 1]{bach2012optimization}). Although this is not the case of neural networks, whose model is highly non-convex due to the presence of the hidden layers, this interpretation does help in visualizing why the resulting formulation provides a group sparse solution. For this reason, we briefly describe it here for the sake of understanding.

\begin{figure*}
	\centering
	\subfloat[$\ell_2$ norm]{\includegraphics[scale=0.85]{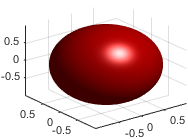}%
		\label{fig:comparison_l2}} %
	\hfil
	\subfloat[$\ell_1$ norm]{\includegraphics[scale=0.85]{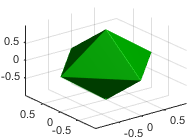}%
			\label{fig:comparison_l1}} %
	\hfil
	\subfloat[$\ell_{2,1}$ norm]{\includegraphics[scale=0.85]{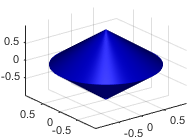}%
			\label{fig:comparison_l21}} %
	\caption{Isosurface for three different regularization terms, with $\mu_{\lambda}=1$. (a) Standard squared $\ell_2$ norm. (b) $\ell_1$ norm enforcing sparsity. (c) $\ell_{2,1}$ norm applied to the groups $\left\{1,2\right\}$ and $\left\{3\right\}$ (without considering the scaling factors).}
	\label{fig:comparison_norms}
\end{figure*}

For a convex loss in \eqref{eq:nn_cost_function}, standard arguments from duality theory show that the problem can be reformulated as follows \cite{ben2001lectures}:
\begin{eqnarray}
\underset{\vect{w}}{\arg\min} & L(\vect{w}) = \displaystyle\frac{1}{N} \sum_{i=1}^N L(\vect{d}_i, \vect{f}(\vect{x}_i)) \nonumber\\
\text{subject to} & R(\vect{w}) \le \mu_\lambda
\label{eq:ivanov_formulation}
\end{eqnarray}
where $\mu_\lambda$ is a scalar whose precise value depends on $\lambda$, and whose existence is guaranteed thanks to the absence of duality gap. In machine learning, this is sometimes called Ivanov regularization, in honor of the Russian mathematician Nikolai V. Ivanov \cite{pelckmans2004morozov}. For a small value of $\mu_\lambda$, such that the constraint in \eqref{eq:ivanov_formulation} is active at the optimum $\vect{w}^*$, it can be shown that the set of points for which $L(\vect{w})$ is equal to $L(\vect{w}^*)$ is tangent to $\mathcal{B} = \left\{ \vect{w} : R(\vect{w}) \le \mu_\lambda \right\}$. Due to this, an empirical way to visualize the behavior of the different penalties is to consider the shape of $\mathcal{B}$ corresponding to them. The shapes corresponding to $\ell_2$ regularization, Lasso, and group Lasso are shown in Fig. \ref{fig:comparison_norms} for a simple problem with three variables. The shape of $\mathcal{B}$ for a weight decay penalty is a sphere (shown in Fig. \ref{fig:comparison_l2}), which does not favor any of the solutions. On the contrary, the Lasso penalty imposes a three-dimensional diamond-shaped surface (shown in Fig. \ref{fig:comparison_l1}), whose vertices lie on the axes and correspond to all the possible combinations of sparse solutions. Finally, consider the shape imposed by the group Lasso penalty (shown in Fig. \ref{fig:comparison_l21}), where we set one group comprising of the first two variables, and another group comprising only the third variable. The shape now has infinitely many singular points, corresponding to solutions having zeroes either on the first and second variables simultaneously, or in the third variable.

\section{Experimental results}
\label{sec:experimental_results}

\subsection{Experimental setup}

In this section, we evaluate our proposal on different classification benchmarks. Particularly, we begin with a simple toy dataset to illustrate its general behavior, and then move on to more elaborate, real-world datasets. In all cases, we use ReLu activation functions \cite{glorot2011deep} for the hidden layers of the network:
\begin{equation}
g_k(s) = \max\left( 0, s \right), \, 1 \le k \le H \,,
\end{equation}
while we use the standard one-hot encoding for the different classes, and a softmax activation function for the output layer. Denoting as $\vect{s}$ the values in input to the softmax, its $i$th output is computed as:
\begin{equation}
g_{H+1}(s_i) = \frac{\exp\left\{ s_i \right\}}{\sum_{j=1}^o \exp\left\{ s_j \right\}} \,.
\end{equation}
The weights of the network are initialized according to the method described in \cite{glorot2010understanding}, and the networks are trained using the popular Adam algorithm \cite{kingma2014adam}, a derivation of stochastic gradient descent with both adaptive step sizes and momentum. In all cases, parameters of the Adam procedure are kept as the default values described in \cite{kingma2014adam}, while the size of the mini-batches is varied depending on the dimensionality of the problem. Specifically, we minimize the loss function in \eqref{eq:nn_cost_function} with the standard cross-entropy loss given by:
\begin{equation}
L(\vect{d}, \vect{f}(\vect{x})) = - \displaystyle\sum_{i=1}^o d_i \log\left( f_i(\vect{x}) \right) \,,
\end{equation}
and multiple choices for the regularization penalty. Dataset loading, preprocessing and splitting is made with the sklearn library \cite{scikit-learn}. First, every input column is normalized in the range $\left[ 0, 1 \right]$ with an affine transformation. Then for every run we randomly keep $25\%$ of the dataset for testing, and we repeat each experiment $25$ times in order to average out statistical variations. For training, we exploit the Lasagne framework,\footnote{\url{https://github.com/Lasagne/Lasagne}} which is built on top of the Theano library \cite{bergstra2010theano}. Open source code to replicate the experiments is available on the web under BSD-2 license.\footnote{\url{https://bitbucket.org/ispamm/group-lasso-deep-networks}}

\subsection{Comparisons with the DIGITS dataset}

\begin{figure*}
	\centering
	\subfloat[Test accuracy]{\includegraphics[scale=0.9]{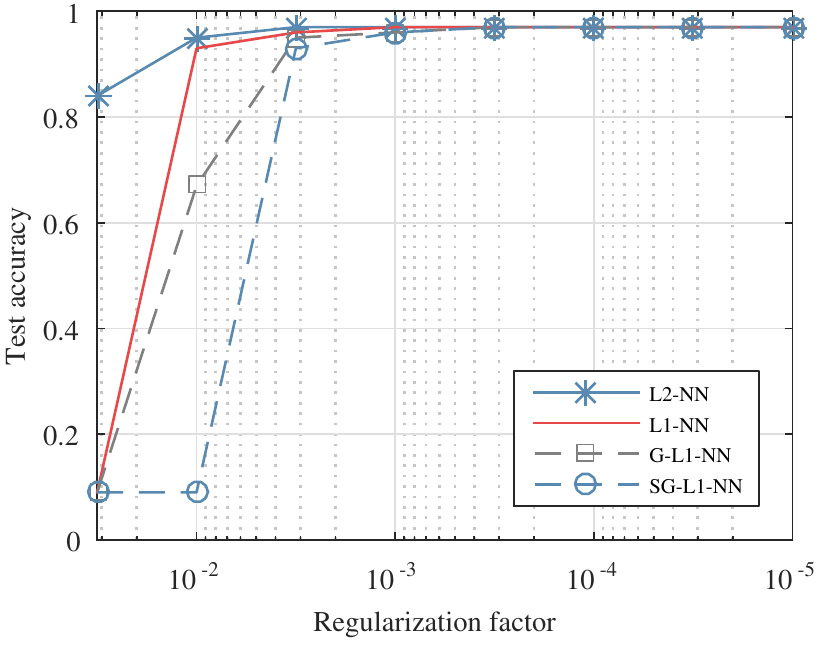}%
		\label{fig:digits_test_accuracy}} %
	\hfil
	\subfloat[Sparsity]{\includegraphics[scale=0.9]{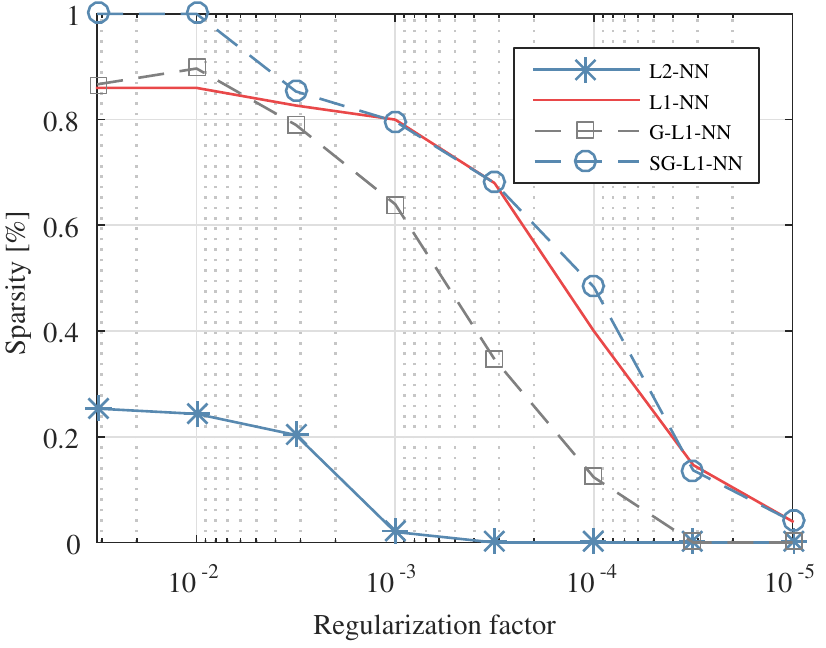}%
		\label{fig:digits_sparsity}} %
	\vfil
	\subfloat[Features selected]{\includegraphics[scale=0.9]{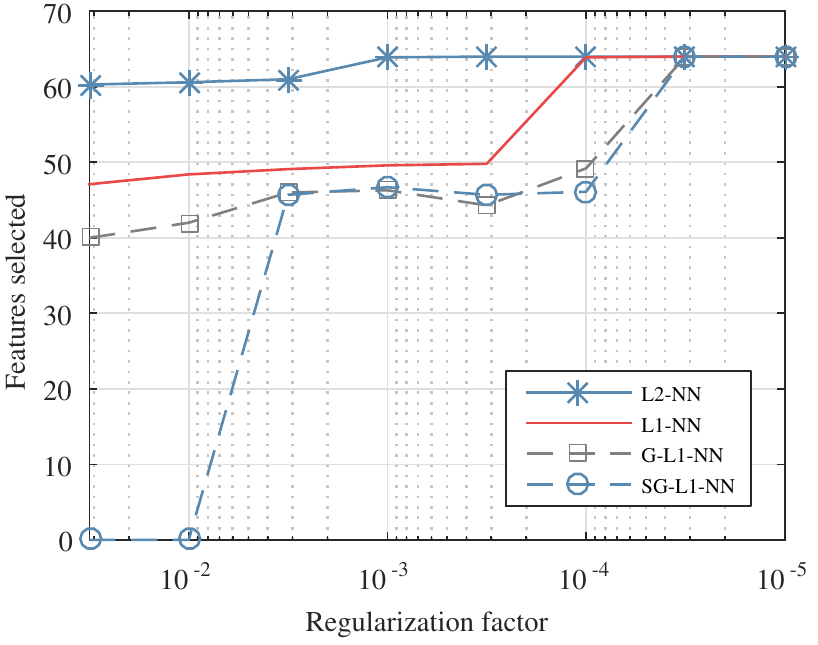}%
		\label{fig:digits_features_selected}} %
	\hfil
	\subfloat[Neurons (hidden layers)]{\includegraphics[scale=0.9]{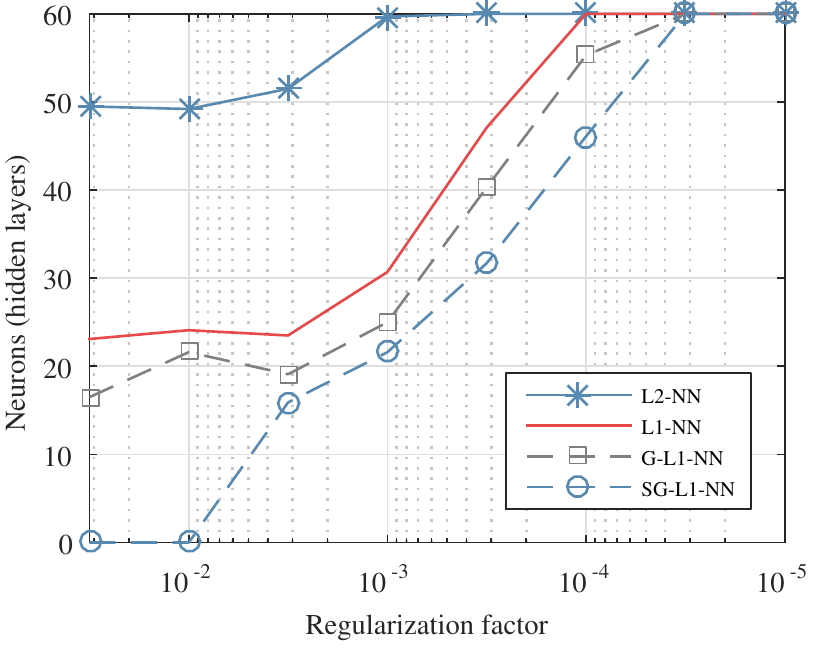}%
			\label{fig:digits_neurons_hidden_layers}} %
	\caption{Results for the digits dataset, when varying the regularization coefficient in $10^{-j}$, $j=1, \ldots, 5$. (a) Test accuracy. (b) Sparsity of the internal connections (in percentage). (c) Number of selected input features. (d) Number of neurons in the hidden layers (total).}
	\label{fig:digits}
\end{figure*}

To begin with, we evaluate our algorithm on a toy dataset of handwritten digit recognition, namely the DIGITS dataset \cite{alimoglu1996methods}. It is composed of $1797$ $8\times8$ grey images of handwritten digits collected from several dozens different people. We compare four neural networks, trained respectively with the weight decay in \eqref{eq:ell_2_regularization} (denoted as L2-NN), the Lasso penalty in \eqref{eq:ell_1_regularization} (denoted as L1-NN), the proposed group Lasso penalty in \eqref{eq:reg_group_lasso} (denoted as G-L1-NN), and finally its sparse variation in \eqref{eq:sgl_regularization} (denoted as SG-L1-NN). In all cases, we use a simple network with two hidden layers having, respectively, $40$ and $20$ neurons. We run the optimization algorithm for $200$ epochs, with mini-batches of $300$ elements. After training, all weights under $10^{-3}$ in absolute value are set to $0$.

\begin{figure}
	\centering
	\subfloat[Example of digit]{\includegraphics[scale=0.5]{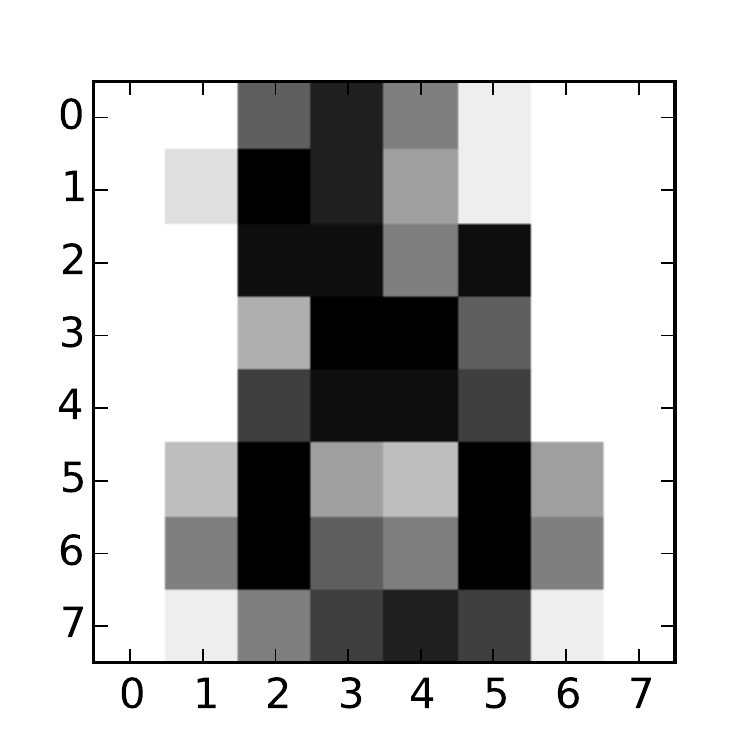}%
		\label{fig:digits_example}} %
	\hfil
	\subfloat[Selected features]{\includegraphics[scale=0.5]{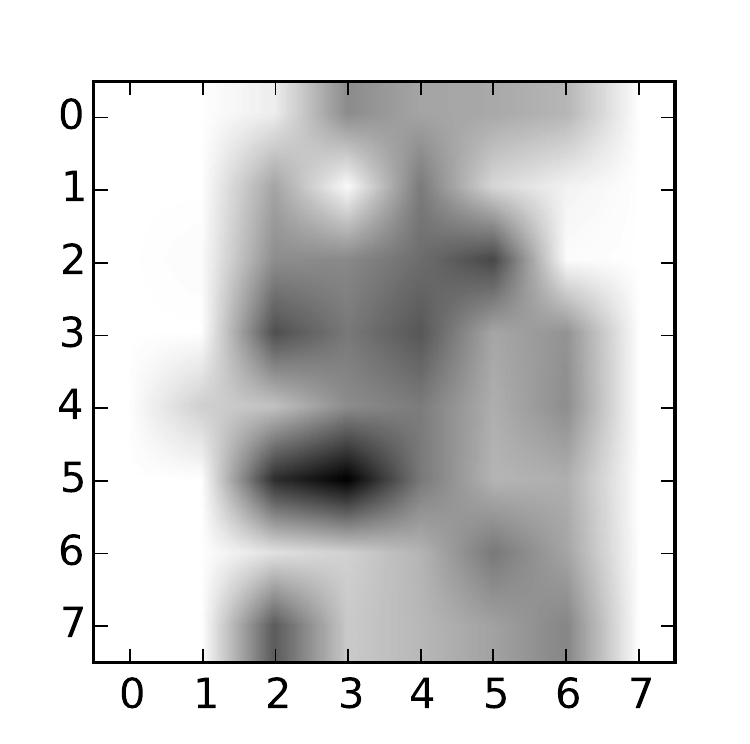}%
		\label{fig:digits_weights_trained}} %
	\caption{Visualization of the selected features for the digits dataset. (a) Example of input pattern to the network (number $8$). (b) Overall strength of outgoing weights from the respective input pixel (white are lowest, black are highest).}
	\label{fig:digits_examples}
\end{figure}

The aim of this preliminary test is to evaluate what we obtain from the different penalties when varying the regularization factor $\lambda$. To this end, we run each algorithm by choosing $\lambda$ in the exponential range $10^{-j}$, with $j$ going from $1$ to $5$. Results of this set of experiments are shown in Fig. \ref{fig:digits}. There are several key observations to be made from the results. To begin with, the overall behavior in terms of test accuracy with respect to the four penalties, shown in Fig. \ref{fig:digits_test_accuracy}, is similar among the algorithms, as they rapidly converge to the optimal accuracy (slightly lower than $100\%$) for sufficiently small regularization factors. In particular, from $10^{-3}$ onwards, their results are basically indistinguishable. Fig. \ref{fig:digits_sparsity} shows the level of sparsity that we obtain, which is evaluated as the percentage of zero weights with respect to the total number of connections. The sparsity of L2-NN is clearly unsatisfactory, oscillating from $20\%$ in the best case to $0\%$ in average. The sparsity of G-L1-NN is lower than the corresponding sparsity of L1-NN, while the results of SG-L1-NN (shown with a dashed blue line) are equal or superior than all alternatives. In particular, for $\lambda = 10^{-3}$ both L1-NN and SG-L1-NN are able to remove four fifths of the connections. At the same time, the resulting sparsity is highly more structured for the proposed algorithm, which is able to consistently remove more features, as shown in Fig. \ref{fig:digits_features_selected}, and neurons in the hidden layers, as shown in Fig. \ref{fig:digits_neurons_hidden_layers}.

Since the input to the classifier is an image, it is quite interesting to visualize which features (corresponding to pixels of the original image) are neglected in the proposed approaches, in order to further validate empirically the proposal. This is shown for one representative run in Fig. \ref{fig:digits_examples}. In Fig. \ref{fig:digits_example} we see a characteristic image in input to the system, representing in this case the number $8$. We see that the digit covers all the image with respect to its height, while there is some white space to its left and right, which is not interesting from a discriminative point of view. In Fig. \ref{fig:digits_weights_trained} we visualize the results of G-L1-NN (which is very similar to SG-L1-NN), by plotting the cumulative intensity of the weights connecting the input layer to the first hidden layer (where white color represents an input with all outgoing connections set to $0$). We see that the algorithm does what we would have expected in this case, by ignoring all pixels corresponding to the outermost left and right regions of the image.

\begin{table*}
\small
\ra{1.3}
\centering
\caption{Schematic description of the datasets.} 
\begin{tabular}{llccclc}
\toprule
Origin & Name & Features & Size & N. Classes & Desired output & Reference \\ \midrule
UCI Repository & Sensorless Drive Diagnosis (SDD) & 48 & 58508 & 11 & Motor operating condition & \cite{bayer2013sensorless} \\ \midrule
MLData Repository & MNIST Handwritten Digits & 784 & 70000 & 10 & Digit (0-9) & \cite{deng2012mnist} \\ \midrule
MLData Repository & Forest Covertypes (COVER) & 54 & 581012 & 7 & Cover type of forest & \cite{blackard1999comparative} \\ \bottomrule
\end{tabular}
\vspace{0.5em}
\label{tab:datasets}
\end{table*}

\subsection{Comparisons with large-scale datasets}

We now evaluate our algorithm on three more realistic datasets, which require the use of deeper, larger networks. A schematic description of them is given in Table \ref{tab:datasets} in terms of features, number of patterns, and number of output classes. The first is downloaded from the UCI repository,\footnote{\url{http://archive.ics.uci.edu/ml/}} while the second and third ones are downloaded from the MLData repository.\footnote{\url{http://mldata.org/}} In the SSD dataset, we wish to predict whether a motor has one or more defective components, starting from a set of $48$ features obtained from the motor's electric drive signals (see \cite{bayer2013sensorless} for details on the feature extraction process). The dataset is composed of $58508$ examples obtained under $11$ different operating conditions. The MNIST database is an extremely well-known database of handwritten digit recognition \cite{deng2012mnist}, composed of $70$ thousands $28 \times 28$ gray images of the digits $0$-$9$. Finally, the COVER dataset is the task of predicting the actual cover type of a forest (e.g. ponderosa pine) from a set of $52$ features extracted from cartographic data (see \cite[Table 1]{blackard1999comparative} for a complete list of cover types). This dataset has roughly a half million training examples, but only $7$ possible classes compared to $11$ and $10$ classes for SSD and MNIST, respectively.

\begin{table}
\small
\ra{1.3}
\centering
\caption{Parameters for the neural networks used in the experiments.} 
\begin{tabular}{lccc}
\toprule
Dataset & Neurons & Regularization & Mini-batch size \\ \midrule
SSD & 40/40/30/11 & $10^{-4}$ & 500 \\ \midrule
MNIST & 400/300/100/10 & $10^{-4}$ & 400 \\ \midrule
COVER & 50/50/20/7 & $10^{-4}$ & 1000 \\ \bottomrule
\end{tabular}
\vspace{0.5em}
\label{tab:network_architectures}
\end{table}

Details on the network's architecture, regularization factor and mini-batch size for the three datasets is given in Table \ref{tab:network_architectures}. Generally speaking, we use the same regularization factor for all algorithms, as it was shown to provide the best results in terms of classification accuracy and sparsity of the network. The network architecture is selected based on an analysis of previous works and is given in the second column of Table \ref{tab:network_architectures}, where $x/y/z$ means a network with one $x$-dimensional hidden layer, a second $y$-dimensional hidden layer, and a $z$-dimensional output layer. We stress that our focus is on comparing the different penalties, and very similar results can be obtained for different choices of the network's architecture and the regularization factors. Additionally, we only consider SG-L1-NN as the previous section has shown that it can consistently outperform the simpler G-L1-NN. 

\definecolor{LightCyan}{rgb}{0.88,1,1}
\begin{table*}
\small
\ra{1.3}
\centering
\caption{Average results on the datasets of Table \ref{tab:datasets}. With a light blue background we highlight the resulting size of the network's layers, including the input one while excluding the softmax one. Times with a $^{\dagger}$ symbol were obtained without using the CUDA backend (see text).} 
\begin{tabular}{lllll}
\toprule
Dataset & Measure & L2-NN & L1-NN & SG-L1-NN \\ \midrule
\multirow{5}{*}{SSD} & Training accuracy [\%] & $0.98$ & $0.99$ & $0.97$ \\
					 & Test accuracy [\%] & $0.98$ & $0.98$ & $0.97$ \\
					 & Training time [secs.] & $445^{\dagger}$ & $496^{\dagger}$ & $416^{\dagger}$ \\
					 & Sparsity [\%] & $[0.17, 0.36, 0.36, 0.16]$ & $[0.51, 0.64, 0.61, 0.43]$ & $[0.64, 0.81, 0.76, 0.54]$  \\
					 & \cellcolor{LightCyan!75}Neurons & \cellcolor{LightCyan!75}$[48.0, 35.5, 24.8, 26.3]$ & \cellcolor{LightCyan!75}$[47.9, 27.5, 19.6, 20.2]$ & \cellcolor{LightCyan!75}$[47.4, 19.0, 14.8, 15.9]$ \\ \midrule
\multirow{5}{*}{MNIST} & Training accuracy [\%] & $0.99$ & $0.99$ & $0.98$ \\
					   & Test accuracy [\%] & $0.98$ & $0.97$ & $0.97$ \\
					   & Training time [secs.] & $81$ & $83$ & $93$ \\
					   & Sparsity [\%] & $[0.60, 0.60, 0.34, 0.08]$ & $[0.91, 0.98, 0.94, 0.44]$ & $[0.96, \approx 1.0, 0.98, 0.48]$ \\
					   & \cellcolor{LightCyan!75}Neurons & \cellcolor{LightCyan!75}$[676.4, 311, 249.9, 93.7]$ & \cellcolor{LightCyan!75}$[658.2, 84.8, 85.1, 73.3]$ & \cellcolor{LightCyan!75}$[581.8, 44.7, 41.0, 60.6]$ \\ \midrule
\multirow{5}{*}{COVER} & Training accuracy [\%] & $0.85$ & $0.84$ & $0.83$ \\ 
					   & Test accuracy [\%] & $0.84$ & $0.83$ & $0.83$ \\
					   & Training time [\%] & $454$ & $479$ & $551$ \\
					   & Sparsity [\%] & $[0.04, 0.10, 0.22, 0.14]$ & $[0.19, 0.48, 0.61, 0.34]$ & $[0.45, 0.82, 0.84, 0.49]$ \\
					   & \cellcolor{LightCyan!75}Neurons & \cellcolor{LightCyan!75}$[54.0, 49.0, 47.3, 18.7]$ & \cellcolor{LightCyan!75}$[53.0, 46.0, 31.0, 14.3]$ & \cellcolor{LightCyan!75}$[52.7, 30.0, 16.0, 11.3]$ \\ \bottomrule
\end{tabular}
\vspace{0.5em}
\label{tab:results}
\end{table*}

\begin{figure}
	\centering
	\subfloat[Example of digit]{\includegraphics[scale=0.5]{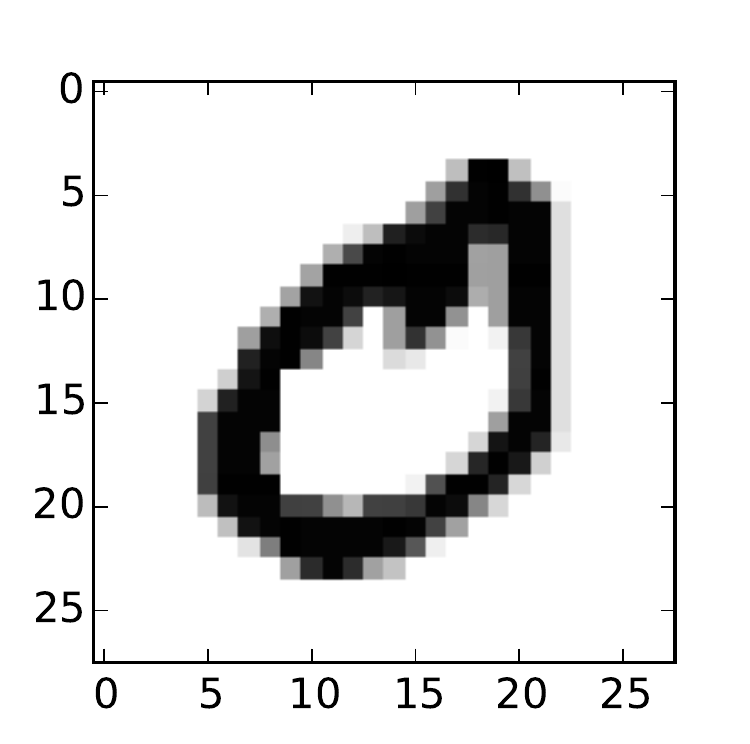}%
		\label{fig:mnist_example}} %
	\hfil
	\subfloat[Selected features]{\includegraphics[scale=0.5]{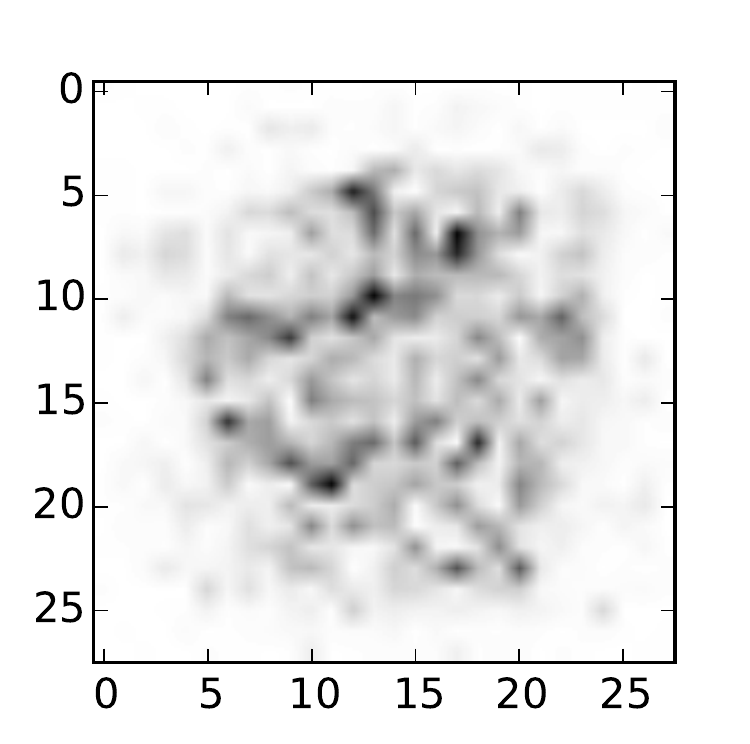}%
		\label{fig:mnist_weights_trained}} %
	\caption{Visualization of the selected features for the MNIST dataset. (a) Example of input pattern to the network (number $0$). (b) Overall strength of outgoing weights from the respective input pixel (white are lowest, black are highest).}
	\label{fig:mnist_examples}
\end{figure}

The results for these experiments are given in Table \ref{tab:results}, where we show the average training and test accuracy, training time, sparsity of the network, and final size of each hidden layer (which is highlighted with a light blue background). As a note on training times, results for the smaller SSD dataset are obtained on an Intel Core i3 @ 3.07 GHz with 4 GB of RAM, while results for MNIST and COVER are obtained on an Intel Xeon E5-2620 @ 2.10 GHz, with 8 GB of RAM and a CUDA back-end employing an Nvidia Tesla K20c. We see that the results in terms of test accuracy are comparable between the three algorithms, with a negligible loss on the MNIST dataset for SG-L1-NN. However, SG-L1-NN results in networks which are extremely sparse and more compact than its two competitors. Let us consider as an example the MNIST dataset. In this case, the algorithm removes more than $200$ features in average from the input vector (compared to approximately $126$ for L1-NN). Also, the resulting network only has $146$ neurons in the hidden layers compared to $243$ for L1-NN and $654$ for L2-NN. Also in this case, we can visually inspect the resulting features selected by the algorithm, which are shown in Fig. \ref{fig:mnist_examples}. In Fig. \ref{fig:mnist_example} we see an example of input pattern (corresponding to the digit $0$), while in Fig. \ref{fig:mnist_weights_trained} we plot the cumulative intensity of the outgoing weights from the input layer. Differently from the DIGITS case, the images in this case have a large white margin on all sides, which is efficiently neglected by the proposed formulation, as shown by the white portions of Fig. \ref{fig:mnist_weights_trained}.

One last observation can be made regarding the required training time. The SGL penalty is actually faster to compute than both the $\ell_2$ and $\ell_1$ norms when the code is run on the CPU, while we obtain a slower training time (albeit by a small margin) when it is executed on the CUDA back-end. The reason for this is the need to compute two square root operations per group in \eqref{eq:sgl_regularization}. This gap can be removed by exploiting several options for faster mathematical computations (at the cost of precision) on the GPU, e.g. by using the `–prec-sqrt' flag on the Nvidia CUDA compiler.

Overall, the results presented in this section show how the sparse group Lasso penalty can easily allow to obtain networks with a high level of sparsity, a low number of neurons (both on the input layer and on the hidden layers), while incurring no or negligible losses in accuracy.

\section{Related works}
\label{sec:related_works}

Before concluding our paper, we describe a few related works that we briefly mentioned in the introduction, in order to highlight some common points and differences. Recently, there has been a sustained interest in methods that randomly decrease the complexity of the network during training. For example, dropout \cite{srivastava2014dropout} randomly removes a set of connections; stochastic depths skips entire layers \cite{huang2016deep}; while \cite{glorot2011deep} introduced the possibility of applying the $\ell_1$ penalty to the activations of the neurons in order to further sparsify its firing patterns. However, these methods are only used to simplify the training phase, while the entire network is needed at the prediction stage. Thus, they are only tangentially related to what we discuss here.

A second class of related works group all the so-called pruning methods, which can be used to simplify the network's structure \textit{after} training is completed. Historically, the most common method to achieve this is the optimal brain damage algorithm introduced by LeCun \cite{lecun1989optimal}, which removes connections by measuring a `saliency' quantity related to the second-order derivatives of the cost function at the optimum. Other methods require instead to compute the sensitivity of the error to the removal of each neuron, in order to choose an optimal subset of neurons to be deleted \cite{suzuki2001simple}. More recently, a two-step learning process introduced by Han \textit{et al.} \cite{han2015learning} has also gained a lot of popularity. In this method, the network is originally trained considering an $\ell_2$ penalty, in order to learn which connections are `important'. Then, the non-important connections, namely all weights under a given threshold, are set to zero, and the network is retrained by keeping fixed the zeroed out weights. This procedure can also be iteratively repeated to further reduce the size of the network. None of these methods, however, satisfy what we considered in the introduction, i.e. they either require a separate pruning process, they do not act directly at the neuron-level, and they might make some heuristic assumptions that should hold at the pruning phase. As an example, the optimal brain damage algorithm is built on the so-called diagonal approximation, stating that the error modification resulting from modifying many weights can be computed by summing the individual contributions from each weight perturbation.

A final class of methods is not interested in learning an optimal topology, insofar as to reduce the actual number of parameters and/or the storage requirements of the network. The most common method in this class is the low-rank approximation method \cite{sainath2013low}, where a weight matrix $\vect{W}_k \in \R^{L_k \times L_{k+1}}$ is replaced by a low-rank factorization $\vect{W}_k = \vect{A}\vect{B}, \, \vect{A} \in \R^{L_k \times r}, \vect{B} \in \R^{r \times L_{k+1}}$, where the rank $r$ must be chosen by the user. Optimization is then performed directly on the two factors instead of the original matrix. The choice of the rank allows to balance between compression and accuracy. As an example, if we wish to compress the network by a factor $p$, we can choose \cite{sainath2013low}:
\begin{equation}
r < \frac{pL_kL_{k+1}}{L_k+L_{k+1}} \,.
\end{equation}
However, this approximation is not guaranteed to work efficiently, and may result in highly worse results for a poor choice of $r$.

\section{Conclusions}
\label{sec:conclusions}

In this paper, we have introduced a way to simultaneously perform pruning and feature selection while optimizing the weights of a neural network. Our sparse group Lasso penalty can be implemented efficiently (and easily) in most software libraries, with a very small overhead with respect to standard $\ell_2$ or $\ell_1$ formulations. At the same time, our experimental comparisons show its superior performance for obtaining highly compact networks, with definite savings in terms of storage requirements and power consumption on embedded devices.

There are two main lines of research that we wish to explore in future contributions. To begin with, there is the problem of studying the interaction between a sparse $\ell_1$ formulation (originated in the case of convex costs), with a non-convex cost as in \eqref{eq:nn_cost_function}, an aspect which is still open in the optimization literature. It would be interesting to investigate the possible improvements with the use of a non-convex sparse regularizer, such as the $\ell_p$ norm with fractional $p$. Alternatively, we might improve the sparse behavior of \eqref{eq:ell_1_regularization} and \eqref{eq:reg_group_lasso} by iteratively solving a \textit{convex} approximation to the original non-convex problem, e.g. by exploiting the techniques presented in \cite{scutari2014decomposition}, as we did in a previous work on semi-supervised support vector machines \cite{scardapane2016distributed}.

Then, we are interested in exploring group Lasso formulations for other types of neural networks, including convolutional neural networks and recurrent neural networks. As an example, we are actively working in extending our previous work on $\ell_1$ sparse regularization in reservoir computing architectures \cite{bianchi2015prediction}, where it is shown that having sparse connectivity can help in creating clusters of neurons resulting in heterogeneous features extracted from the recurrent layer.

\section*{Acknowledgments}

The authors wish to thank Dr. Paolo Di Lorenzo for his insightful comments.

\bibliographystyle{IEEEtran}
\bibliography{biblio}

\end{document}